\documentclass{article}
\usepackage{spconf,amsmath,epsfig}

\usepackage[utf8]{inputenc} %
\usepackage[T1]{fontenc}    %

\usepackage[english]{babel}

\usepackage{microtype}      %
\usepackage{xspace}
\usepackage{csquotes}
\usepackage[normalem]{ulem}
\usepackage[detect-all, binary-units]{siunitx}

\usepackage{amsfonts,amsmath}       %
\usepackage{mathtools}

\usepackage{xspace}
\usepackage{url}
\usepackage{array,booktabs}
\usepackage{subcaption}
\usepackage{enumitem}
\setlist{nosep}
\usepackage[pdftex=true,breaklinks=true,hidelinks=true,colorlinks=true,citecolor=blue]{hyperref}
\makeatletter
\define@key{href}{font}{#1}
\makeatother
\usepackage{xpatch}
\newcommand\hrefdefaultfont{\ttfamily\small}
\xpatchcmd\href{\setkeys{href}{#1}}{\setkeys{href}{font=\hrefdefaultfont,#1}}{}{\fail}

\usepackage[
  backend=biber,
  citestyle=numeric-comp,
  bibstyle=ieee,
  sorting=nyt,
  backref=false,
  dashed=false,
  alldates=year,
  maxcitenames=2,
  mincitenames=1,
  url=false,
  doi=true,
  eprint=false,
]{biblatex}

\DeclareNameAlias{sortname}{last-first}
\AtBeginBibliography{\ninept}
\DeclareSourcemap{
  \maps[datatype=bibtex, overwrite]{
    \map{
      \step[fieldset=editor, null]
    }
  }
}

\usepackage[
  capitalize,
]{cleveref}

\usepackage[dvipsnames,svgnames]{xcolor}
\usepackage{graphicx}
\usepackage[
  inkscape=true, %
  inkscapearea=crop,
]{svg}
\usepackage{placeins}
\usepackage[
  acronym,
]{glossaries-extra}

\setabbreviationstyle[acronym]{long-short}
\defglsentryfmt{\ifglsused{\glslabel}{\glsgenentryfmt}{\textit{\glsgenentryfmt}}}
\defglsentryfmt[acronym]{\ifglsused{\glslabel}{\glsgenentryfmt}{\textit{\glsgenentryfmt}}}
\glsdisablehyper

\newacronym{gl:hcat}{HCAT}{hierarchical crop and agriculture taxonomy}
\newacronym{gl:hcatv2}{HCATv2}{hierarchical crop and agriculture taxonomy -- version 2}
\newacronym{gl:cap}{CAP}{common agricultural policy}
\newacronym{gl:eu}{EU}{European Union}
\newglossaryentry{gl:admindata}{name=administrative data, description={}}
\newglossaryentry{gl:govdata}{name=government data, description={}}
\newglossaryentry{gl:phenology}{name=phenology, description={}}
\usepackage{xargs}

\newcommand{\eurocrops}{\textsc{EuroCrops}\xspace}
\newcommand{\tinyeurocrops}{\textsc{TinyEuroCrops}\xspace}

\newcommand{\github}{\textsc{GitHub}\xspace}

\newcommand{\class}[1]{\textsl{#1}}
\newcommand{\country}[1]{\emph{#1}}

\newcommand{\brand}[1]{\textsc{#1}\xspace}

\newcommand{\project}[1]{\textsl{#1}\xspace}

\newcommand{\HCATNum}[1]{\num[minimum-integer-digits=2,mode=text]{#1}}
\newcommandx{\HCAT}[5][2=,3=,4=,5=]{%
  {\ttfamily%
    \HCATNum{#1}%
    \notblank{#2}{--\HCATNum{#2}}{}%
    \notblank{#3}{--\HCATNum{#3}}{}%
    \notblank{#4}{--\HCATNum{#4}}{}%
    \notblank{#5}{--\HCATNum{#5}}{}%
  }%
}

\usepackage{tumcolors}
\usepackage{tumabbrev}

\usepackage[
  colorinlistoftodos,
  color={red!100!green!33},
  figwidth=.5\linewidth,
]{todonotes}

\presetkeys{todonotes}{%
  size=\tiny,
}{}

\usepackage{setspace}
\newcounter{mycomment}
\newcommandx{\mynote}[3][
  1=, %
  3=, %
]{%
  \refstepcounter{mycomment}{%
    \setstretch{0.7}%
    \todo[#3]{\textbf{#1\themycomment:} #2}%
  }%
}

\newcommand{\reduline}{\bgroup {\textcolor{red}\ULdepth=-.55ex} \ULset}

\title{%
  Harnessing Administrative Data Inventories to Create a Reliable Transnational Reference Database for Crop Type Monitoring%
}
\name{%
  Maja Schneider%
  \textsuperscript{*}%
  \thanks{%
    \textsuperscript{*}corresponding author\newline\footnotesize
    M. Schneider, M. Körner, and \eurocrops receive funding from the German \emph{Federal Ministry for Economic Affairs and Climate Action} on the basis of a resolution of the German Bundestag under reference {\sffamily\scriptsize 50EE1908} and from the European Union's \emph{Horizon 2020} research and innovation programme under grant agreement No {\sffamily\scriptsize 101004112}.
    The authors thank the \emph{Stifterverband} for supporting \eurocrops with the Open Data Impact Award 2021.
  }%
  and
  Marco Körner
}
\address{%
  Technical University of Munich (TUM),
  TUM School of Engineering and Design,\\
  Department of Aerospace and Geodesy,
  Arcisstr.~21,
  80333~Munich, Germany\\
  \href{mailto:maja.schneider@tum.de}{\{maja.schneider}, \href{mailto:marco.koerner@tum.de}{marco.koerner\}}\href{mailto:maja.schneider@tum.de}{@tum.de}
}

\graphicspath{{./images/}}

\newcommand{\figCountries}{%
  \begin{figure}[t]
    \centering
    \includesvg[width=\linewidth]{countries}
    \caption{
      \eurocrops compiles crop type reference data from several \emph{\glsfirst{gl:eu}} countries.
      This data has been extracted from self-declarations that farmers submit to receive subsidies under the \emph{\glsfirst{gl:cap}}.
      Several countries, shaded in \textcolor{tumgreen}{green}, already publish this information, 
      in contrast to those shaded in \textcolor{tumorange}{orange}.
    }
    \label{fig:countries}
  \end{figure}
}

\newcommand{\figEuroCropsTimeline}{%
  \begin{figure}[t]
    \centering
    \includesvg[width=\linewidth, pretex=\tiny]{eurocrops_timeline.svg}
    \caption{
    	The work on \eurocrops project is scheduled for the next years until 2024.
      Beyond the primary data collection activities carried out continuously during the entire project time span, we plan to use \eurocrops as a solid basis for creating teaching and learning concepts and to build reliable benchmarking tools that will be offered to the public.
    }
    \label{fig:timeline}
  \end{figure}
}

\newcommand{\figPies}{%
  \begin{figure}[t]
    \centering
    \includegraphics[width=.8\linewidth]{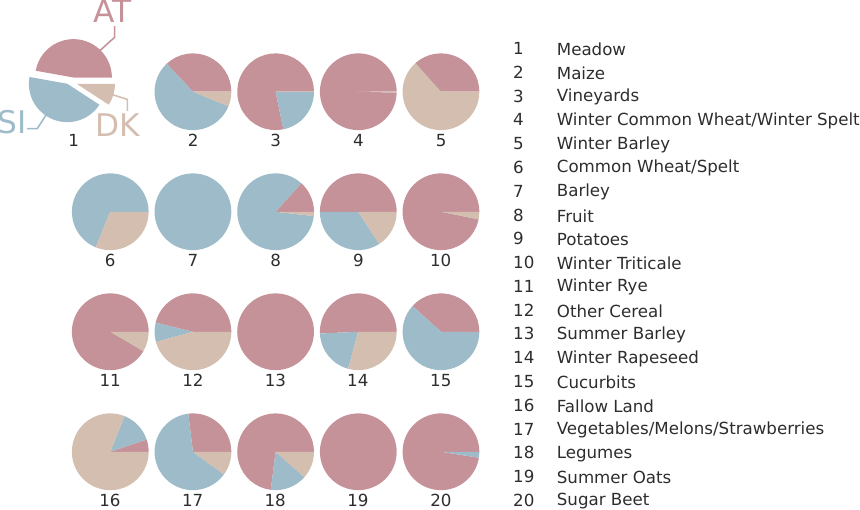}
    \caption{
      \eurocrops allows for analysing the statistics of cultivated crops and their comparison across the different contributing countries.
      This information can be used to derive biodiversity indicators.
    }
    \label{fig:pies}
  \end{figure}
}

\newcommand{\figTaxonomy}{%
\begin{figure}[t]
  \centering
  \includesvg[width=.8\linewidth]{images/tax_with_labels}
  \caption{%
    The \emph{\glsfirst{gl:hcat}}~\cite{Schneider21:EPE} uniquely encodes each crop species and allows to compare self-declaration data transnationally across countries.
    Note that this visualization has been simplified for visualization purposes.
  }
  \label{fig:tax}
\end{figure}
}

\newcommand{\figParcels}{%
\begin{figure}[t]
  \centering
  \includesvg[width=.8\linewidth]{parcels}
  \caption{
    \eurocrops compiles precise geometric representations of each field parcel and enriches these with their associated \gls{gl:hcat}-encoded crop class label.
  }
  \label{fig:parcels}
\end{figure}
}
\newcolumntype{x}{l}
\newcolumntype{X}{>{\scriptsize}l}
\newcolumntype{y}{r}
\newcolumntype{Y}{>{\scriptsize}r}
\newcolumntype{v}[1]{>{\raggedright\arraybackslash\hspace{0pt}}p{#1}}
\newcolumntype{V}[1]{>{\scriptsize\raggedright\arraybackslash\hspace{0pt}}p{#1}}

\begin{document}
\maketitle
\begin{refsection}%
\tikz[remember picture, overlay] {%
  \node[draw=tumivory, anchor=north, below=of current page.north, text width=.825\paperwidth, inner sep=.5em, font=\small] {%
    This paper was originally published as:\newline
    \fullcite*{Schneider22:HAD}
  };
}%
\end{refsection}
\begin{abstract}
  With leaps in machine learning techniques and their application on Earth observation challenges has unlocked unprecedented performance across the domain.
  While the further development of these methods was previously limited by the availability and volume of sensor data and computing resources, the lack of adequate reference data is now constituting new bottlenecks.
  Since creating such ground-truth information is an expensive and error-prone task, new ways must be devised to source reliable, high-quality reference data on large scales.
  As an example, we showcase \eurocrops, a reference dataset for crop type classification that aggregates and harmonizes administrative data surveyed in different countries with the goal of transnational interoperability.
\end{abstract}
\section{Introduction}
\label{sec:introduction}

In recent years, data-driven methods addressing remote sensing and Earth observation problems have shown impressive performance~\parencite{Dubovik21:GCR}.
While the development of such approaches used to be limited by the amount of available observational data and computing resources, these determining factors have now vanished.
Modern Earth observation programs provide a multitude of data products across manifold spectral, spatial, and temporal resolutions with today's data processing pipelines able to process these data volumes~\parencite{Sudmanns20:BED}.
These have been used to compile various general-purpose~\parencite{Schmitt19:SCD} or application-specific datasets~\parencite{Russwurm20:BTS,Turkoglu21:CMI}.
Instead, the lack of appropriate reference data---\ie \emph{labels}, \emph{annotations}, or \emph{targets}---is now the new bottleneck that restricts the further development of data-driven modelling and information extraction techniques.
To keep pace with the ever-growing and expanding data archives, the properties of reference data must align with those of the Earth observation data.
This, in particular, calls for improved quantity, quality, resolution, and temporal frequency of reference data.
However, this is difficult to achieve with established annotation processes, \eg manual labelling procedures or iterative and interactive curation protocols.

To address this problem, we want to motivate the use of \gls{gl:admindata} and describe its far-reaching possibilities.
With the example of the \eurocrops initiative, we demonstrate how pre-existing metadata can be used to derive reliable, high-quality and interoperable reference datasets on large spatial and temporal scales.

\glsreset{gl:admindata}
\section{Administrative Data}

With an evergrowing focus on data production and collection across society,
the volume of
\gls{gl:admindata} or \gls{gl:govdata}
has also been increasing exponentially in recent years \parencite{Reinsel21:WGD}.
This data is acquired, collected, or compiled by public authorities to support and enable government services and processes, including planning or monitoring, and thus serves as the basis for decisions and interventions.

To serve these purposes, \gls{gl:admindata} is usually collected in a continuous and regular manner.
These collections are characterised by high granularity and wide coverage, must meet strict quality as well as reliability standards, and need to be well documented.
\Gls{gl:admindata} may contain items that provide unique keys across different sources, \eg geocodes, instance identifiers, acronyms or pseudonyms, or further properties.
Despite all of this, the use of \gls{gl:admindata} is complex and difficult.
As a consequence of their distributed acquisition, data is usually available in different, often proprietary, formats or lacking any standardisation, limiting its use and interoperability.
Likewise, incompatible formats require prior harmonisation, alignment, and aggregation or disaggregation \parencite{Marini20:BDE}.
The fact that \gls{gl:admindata} is surveyed from different contexts and organisational levels---\eg on supra-national, trans-/international, national, sub-national, (inter-)regional, or at community level---constitutes another impediment to its further use.
Finally, there is a lack of central and global repositories, obscuring the existence of datasets to scientists and engineers, while their use is further restricted by privacy protection and data agency legislation, their closed-source nature and their huge volumes of data.

Considering the aforementioned observations, the use of \gls{gl:admindata} remains to offer enormous potential \parencite{Connelly16:RAD}.
Being able to access and derive datasets from them allows for cross-validation to assess and improve the quality of the data.
The wide coverage of continuously maintained \gls{gl:admindata} enables both spatial and temporal analysis to be conducted.
Encouragingly, ever more countries are developing their open data strategies \parencite{%
  Quarati21:OGD%
}, foreshadowing that \gls{gl:admindata} will be widely and affordably available in the not too distant future.
\section{The \eurocrops Project}
\figCountries

These considerations
discussed so far
speak of the difficulties of using \gls{gl:admindata} for scientific purposes.
To counter this, we initiated the \eurocrops project, where we aggregate and harmonise administrative data surveyed in different countries of the \gls{gl:eu}, with the goal of transnational interoperability.

\subsection{Motivation and Idea}

Instantiating a motivating use-case, we have chosen the problem of crop type classification from optical remote sensing imagery.
This is an inherently difficult task, as vegetation features remarkable biological and geographic diversity and, therefore, requires high-capacity approaches to be modelled in a data-driven way.
Further, vegetation processes are intrinsically time-dependent, \ie temporal sequences of observations are required to capture the underlying dynamics appropriately.
Fortunately, modern Earth observation satellites provide this information in abundance, and current data-driven machine learning approaches are showing unprecedented performance in predicting and classifying plant species \parencite{Russwurm20:SAR,Russwurm18:MTL,Turkoglu21:CMI,Schneider21:SIT}.
Nevertheless, these approaches are limited by the amount and quality of the corresponding ground-truth information.

\subsection{Database}
\figParcels

Manually annotating Earth observation data with the particular cultivated plant species grown on each field parcel (\cf \cref{fig:parcels}) becomes infeasible on large scales; new ways of acquiring this information are in need.
For \eurocrops, we take advantage of the \gls{gl:cap} set up by the \gls{gl:eu}.
This requires the farmers of each member state to declare the crop species cultivated on their parcels in order to receive subsidies accordingly.
Although data is collected and archived by the authorities of the respective state, the majority of them keep it undisclosed.

\eurocrops aims at compiling and harmonising such data to demonstrate its widespread potential.

\figTaxonomy
\subsection{Taxonomy}

\eurocrops is intended to be used in combination with any kind of remote sensing data and transnationally across spatial scales.
As the various country-specific protocols to represent crop species are typically mutually incompatible, it was necessary to
harmonise them appropriately.
This gave rise to the \gls{gl:hcat} \parencite{Schneider21:EPE}, extending the existing \emph{EAGLE} matrix \parencite{Arnold13:EAGLE}.
As visualized in \cref{fig:tax}, each crop species can be uniquely encoded with a specific multi-level \gls{gl:hcat} identifier
\begin{equation*}
  \underbracket{[33]}_{\substack{\text{position in}\\ \text{\emph{EAGLE} matrix}}}\text{--}
  \underbracket{\text{XX}}_\text{Level 3}\text{--}
  \underbracket{\text{XX}}_\text{Level 4}\text{--}
  \underbracket{\text{XX}}_\text{Level 5}\text{--}
  \underbracket{\text{XX}}_\text{Level 6}
\end{equation*}
ranging from
\class{Cereals} as \HCAT{33}[1][1][0][0] (level 4)
to
\class{Summer Oats} as \HCAT{33}[1][1][5][3] (level 6).
This representation scheme allows for comparing crop species across contributing \gls{gl:eu} countries.

\subsection{Fields of Application}

\figPies

\eurocrops is data-agnostic and designed to be used together with any kind of geo-referenced Earth observation data.
Thus, the field of possible applications is vast and ample.
By addressing the motivating use-case of crop type classification using this reference dataset, data-driven methods will implicitly learn internal representations of vegetation dynamics, \ie its \gls{gl:phenology}.
Such models can further be used to monitor the development of vegetation stocks to spot unexpected patterns, \eg caused by environmental influences \parencite{Marszalek22:ECT}, or to predict the expected yield  \parencite{Marszalek22:PMY}.
Furthermore, the country-specific statistics compiled and harmonised in \eurocrops allow for statistical investigation of regional distributions of crop cultivation patterns (\cf \cref{fig:pies}) and to derive biodiversity markers at various regional scales~\cite{Schneider22:HHC}.
Owing to its broad applicability, \eurocrops provides the basis for further research projects.%
\footnote{%
  \eg
  \project{Global Earth Monitor} (\href[font=\footnotesize\ttfamily]{https://www.globalearthmonitor.eu/}{www.globalearthmonitor.eu}),
  \project{PreTrainAppEO} (\href[font=\footnotesize\ttfamily]{https://www.asg.ed.tum.de/lmf/pretrainappeo/}{www.asg.ed.tum.de/lmf/pretrainappeo}),
  \project{DUKE} (\href[font=\footnotesize\ttfamily]{https://www.asg.ed.tum.de/lmf/duke/}{www.asg.ed.tum.de/lmf/duke}),
  \etc
}

\subsection{Current State}

Since the start of the project, we collected declaration data from 16 countries across the years 2015 to 2021 (\cf \cref{fig:countries}) and harmonised one year per country.
To demonstrate the use of \eurocrops, we compiled the early demonstrator dataset \tinyeurocrops~\parencite{Schneider21:TEC} containing reference data from \country{Austria}, \country{Germany}, and \country{Slovenia}.
The current state of \eurocrops can be tracked through the project website \href{https://www.eurocrops.tum.de}{www.eurocrops.tum.de} and the associated \github repository \href{https://github.com/maja601/EuroCrops}{github.com/maja601/EuroCrops}.
The latter also serves as a platform for tutorials, ongoing discussions, requests, and bug tracking.

\subsection{Plans and Outlook}

\figEuroCropsTimeline

With \eurocrops, we not only want to provide the reference dataset described hitherto but rather to demonstrate its far-reaching potential on a broad scale.
As visualised in \cref{fig:timeline}, the project envisages several further initiatives:
\smallskip
\begin{description}[style=unboxed, leftmargin=0cm]
  \item[Annual Curation and Extension]
  The \eurocrops project schedule foresees actions until the year 2024.
  Within that time, continuous updates and temporal extensions to the the existing dataset will be applied with new reference data.
  Simultaneously, more countries will be added, as they volunteer to contribute their data.
  \smallskip
  \item[Tutorials]
  We intend to use the \eurocrops dataset as a sound foundation to introduce further user groups to the emerging topics of machine learning in the context of Earth observation.
  For this purpose, teaching and learning modules for self-study, as well as for the use in interactive tutorial formats, will be developed.
  These workshops will be held at upcoming community events, conferences and workshops.
  \smallskip
  \item[Benchmarks and Challenges]
  In order to assist researchers and users in the evaluation of their data processing pipelines, we will further provide tools and methods to benchmark several tasks related to crop type classification using regionally and temporally held-out subsets of \eurocrops.
\end{description}

\section{Summary}

The increase of remote sensing data and compute resources within recent years no longer restrict the development of data-driven Earth observations models.
Instead, the community is now lacking reference data matching the properties of the data in remote sensing archives.
To counteract this new bottleneck, we presented \eurocrops to showcase how existing \gls{gl:admindata} can be used to derive high-quality, reliable ground-truth annotations to train and evaluate data-driven remote sensing and Earth observation models.
With its analysis-ready design and widespread availability, \eurocrops also targets experts from other research communities and invites them to address prevalent problems across remote sensing analysis and Earth observation research.

\eurocrops will be made available through several remote sensing data repositories, e.g., \brand{GeoDB}, \brand{CODE-
DE}, \brand{eoLearn}, and \brand{Google Earth Engine}.

\section{References}
\label{sec:ref}

\setlength{\bibitemsep}{1pt}
\defbibenvironment{bibliography}
{\list
  {\printtext[labelnumberwidth]{%
      \printfield{labelprefix}%
      \printfield{labelnumber}}}
  {\setlength{\labelwidth}{\labelnumberwidth}%
    \setlength{\leftmargin}{\labelwidth}%
    \setlength{\labelsep}{\biblabelsep}%
    \addtolength{\leftmargin}{\leftskip}%
    \setlength{\itemsep}{\bibitemsep}%
    \setlength{\parsep}{\bibparsep}}%
  \renewcommand*{\makelabel}[1]{\hss##1}}
{\endlist}
{\item}

\renewcommand{\UrlFont}{\ttfamily\scriptsize}
\printbibliography[heading=none]

\end{document}